\documentclass[sigconf,screen,nonacm]{acmart}

\AtBeginDocument{%
  \providecommand\BibTeX{{%
    \normalfont B\kern-0.5em{\scshape i\kern-0.25em b}\kern-0.8em\TeX}}}

\copyrightyear{2025}
\acmYear{2025}
\setcopyright{rightsretained}

\usepackage{multirow}
\usepackage{makecell}
\usepackage{caption}

\begin{document}

\title[CLIP the Landscape]{CLIP the Landscape: Automated Tagging of Crowdsourced Landscape Images}

\author{Ilya Ilyankou}
\authornote{Both authors contributed equally to this research}
\email{ilya.ilyankou.23@ucl.ac.uk}
\orcid{0009-0008-7082-7122}

\author{Natchapon Jongwiriyanurak}
\authornotemark[1]
\email{natchapon.jongwiriyanurak.20@ucl.ac.uk}
\orcid{0000-0002-6188-8432}

\affiliation{%
  \institution{UCL SpaceTimeLab}
  \city{London}
  \country{UK}
}

\author{Tao Cheng}
\email{tao.cheng@ucl.ac.uk}
\orcid{0000-0002-5503-9813}

\author{James Haworth}
\email{j.haworth@ucl.ac.uk}
\orcid{0000-0001-9506-4266}

\affiliation{%
  \institution{UCL SpaceTimeLab}
  \city{London}
  \country{UK}
}


\begin{abstract}
We present a CLIP-based, multi-modal, multi-label classifier for predicting geographical context tags from landscape photos in the Geograph dataset--a crowdsourced image archive spanning the British Isles, including remote regions lacking POIs and street-level imagery. Our approach addresses a Kaggle competition \cite{kaggle} task based on a subset of Geograph's 8M images, with strict evaluation: exact match accuracy is required across 49 possible tags. We show that combining location and title embeddings with image features improves accuracy over using image embeddings alone. We release a lightweight pipeline\footnote{https://github.com/SpaceTimeLab/ClipTheLandscape} that trains on a modest laptop, using pre-trained CLIP image and text embeddings and a simple classification head. Predicted tags can support downstream tasks such as building location embedders for GeoAI applications, enriching spatial understanding in data-sparse regions.
\end{abstract}



\received{29 May 2025}

\maketitle

\captionsetup{skip=0.1pt} 

\section{Introduction}

\begin{figure}[ht]
    \centering
    \includegraphics[width=\linewidth]{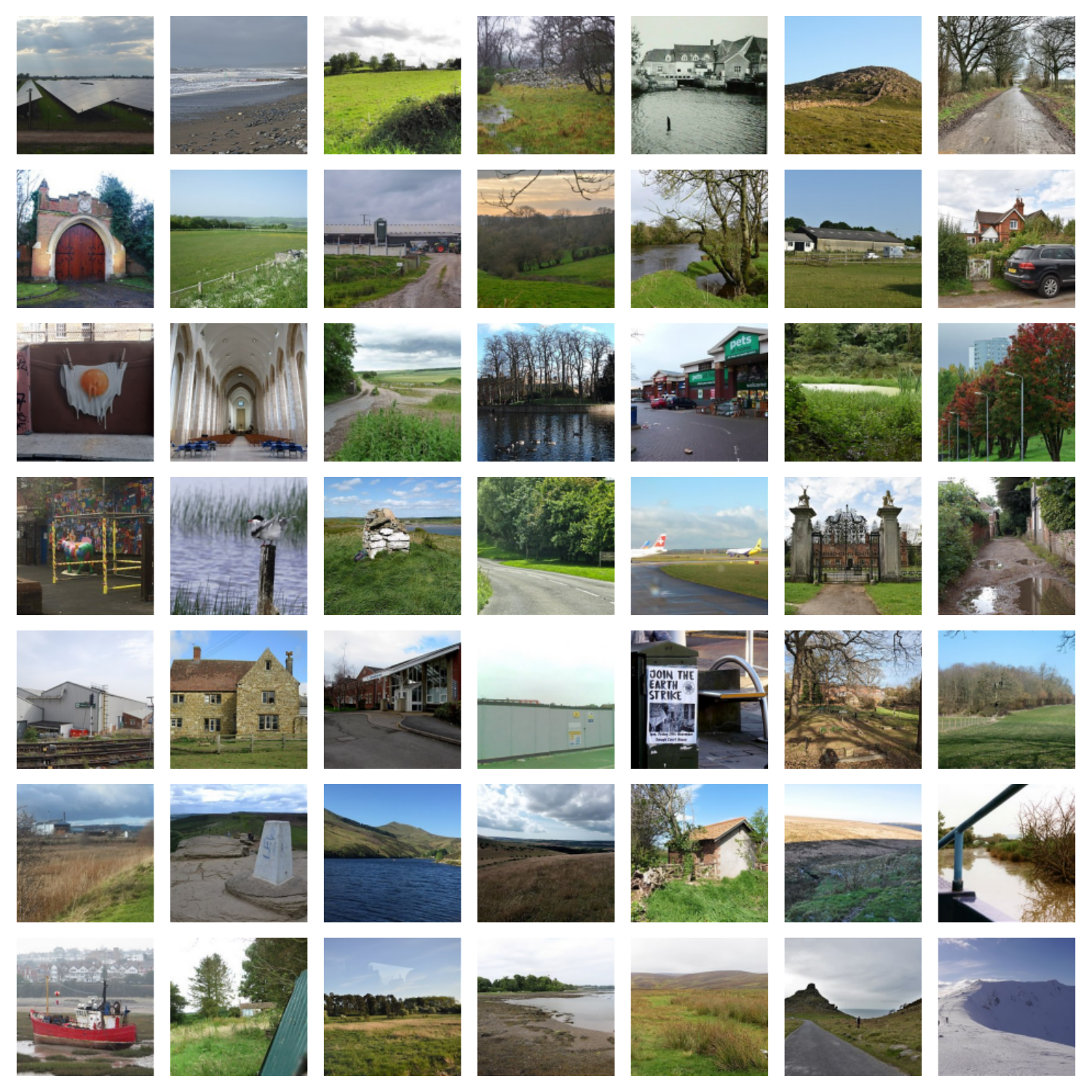}
    \caption{Example training set images for all 49 tags.} 
    \label{fig:7x7-collage}
\end{figure}

We address the Kaggle competition \cite{kaggle} launched by the Geograph project\footnote{\url{https://www.geograph.org.uk/}}, which focuses on multi-label classification of user-contributed landscape images. Geograph is a crowdsourced platform hosting over 8 million publicly available images of Britain and Ireland, each accompanied by metadata. While recent uploads include taxonomy tags describing image content and location, older images (submitted before tagging was mandatory) remain unlabelled and are thus harder to discover and retrieve from the website. This motivates the development of automated methods to predict these content and location-based tags. Figure~\ref{fig:7x7-collage} shows one example image per tag from the training set, illustrating the diversity across the 49 most popular taxonomy categories (see Table~\ref{tab:modalities} in the Appendix for the tag list).

Incomplete or incorrect annotations are a common issue in crowdsourced multi-label datasets \cite{lin_rethinking_2022}, particularly in landscape classification, where user interpretations vary and subjectivity is high \cite{callau_landscape_2019}. This makes landscape image classification particularly challenging compared to standard computer vision tasks with more objective ground truth.

We address these challenges by leveraging the rich multimodal information in the Geograph dataset and using CLIP, a Transformer-based vision–language model trained to learn joint image–text embeddings via contrastive learning. Its speed, flexibility, and zero-shot capabilities have made it popular in geospatial applications, including walkability analysis \cite{liu_new_2023}, remote sensing scene classification \cite{li_rs-clip_2023}, street safety \cite{jongwiriyanurak_et_al:LIPIcs.GIScience.2023.44,wang-clip-safety} and urban function analysis \cite{huang_zero-shot_2024}.

In this work, we use frozen embeddings from a pre-trained \texttt{ViT-B/32} CLIP model and train lightweight Linear and Multilayer perceptron (MLP) classification heads with and without MixUp regularisation \cite{zhang_mixup_2018}. The models are trained on over half a million Geograph images using different combinations of input modalities: image features, text (image title), and approximate location. Since this task involves predicting human tagging behaviour, we find that the image title alone is often a stronger predictor of tags than the image itself. However, the best results are achieved when all three modalities are fused.

Our contributions include demonstrating the effectiveness of multimodal fusion for crowdsourced image tagging and providing insights into the relative importance of different modalities for landscape classification.

\section{Data and Method}

\subsection{Dataset}

\begin{figure}[ht]
    \centering
    \includegraphics[width=\linewidth]{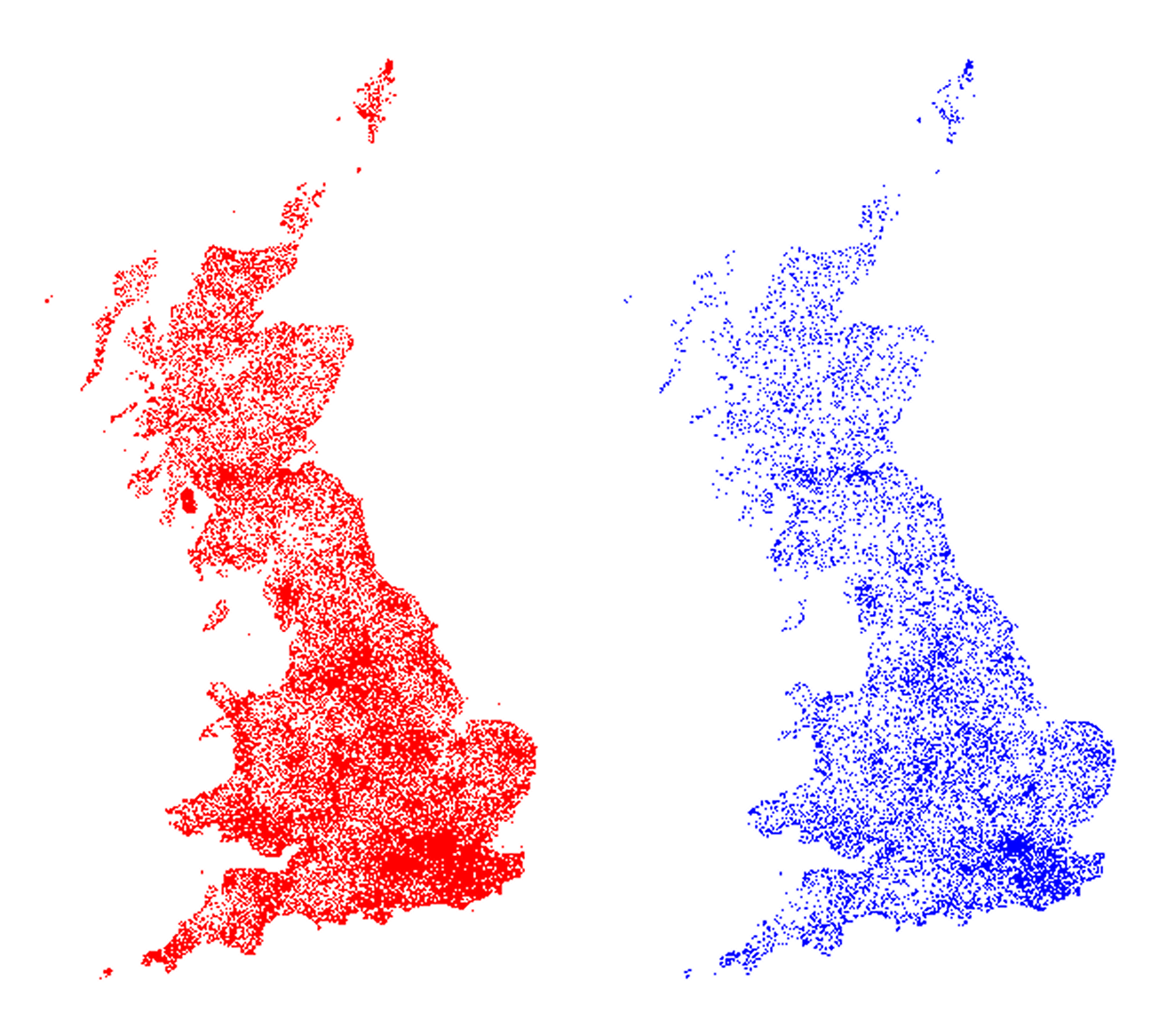}
    \caption{National grid cells in Britain containing at least one image in the training (red) or test (blue) set.}
    \label{fig:uk-train-test-distribution}
\end{figure}

As of May 2025, the Geograph platform hosted 8.02M photographs covering most of the UK and, to a lesser extent, Ireland, released under the CC BY-SA 4.0 license. The Kaggle challenge provides 657k images for training and 135k for testing and final evaluation, released under the CC BY-SA 2.0 license. All images are resized to $224 \times 224$ pixels and include metadata such as contributor ID and name, image title, and the National Grid reference ($1 \times 1km$). Training and test images are well-distributed across Britain, as shown in Figure \ref{fig:uk-train-test-distribution}.


\begin{figure}[ht]
    \centering
    \includegraphics[width=\linewidth]{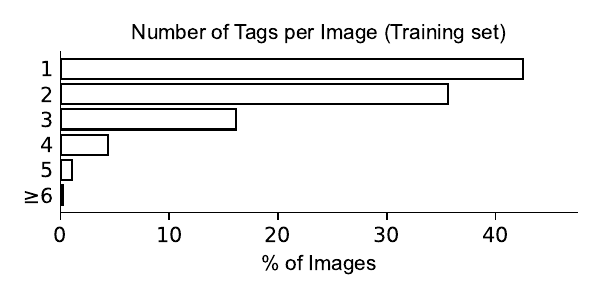}
    \caption{Tag distribution across the training set images.}
    \label{fig:tag-distribution}
\end{figure}

Training images are annotated with one or more of 49 popular tags (multi-hot encoded), such as `Coastal', `Public buildings and spaces', or `Wild Animals, Plants and Mushrooms'. Tag names and descriptions are listed in the Appendix. The majority of training images have between 1--3 tags, as shown in Figure \ref{fig:tag-distribution}.


\begin{figure}[ht]
    \centering
    \includegraphics[width=\linewidth]{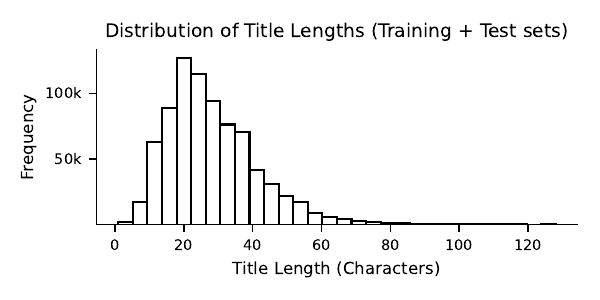}
    \caption{Distribution of title lengths.}
    \label{fig:title-length-distribution}
\end{figure}

\begin{figure}[ht]
    \centering
    \includegraphics[width=1\linewidth]{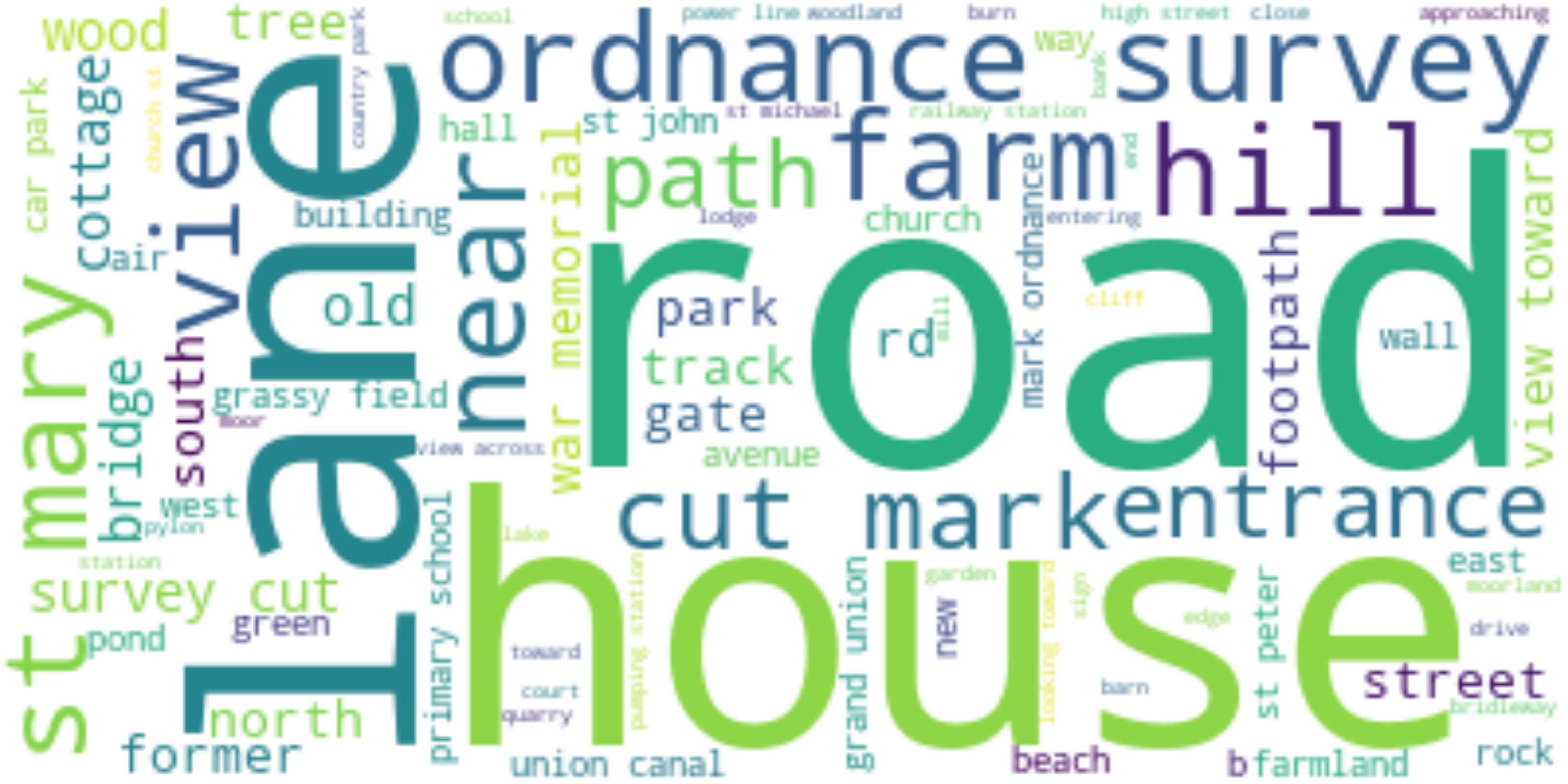}
    \caption{Common words in image titles.}
    \label{fig:wordcloud}
\end{figure}

Image titles are available for 99.99\% of the dataset. They average 28 characters in length, ranging from generic (\emph{`Lying Snow', `Fire!'}) to highly descriptive (\emph{`View of the Shard, Walkie Talkie, Gherkin, Tower 42 and Heron Tower from a train from Hertford East approaching Liverpool Street'}). The longest titles are 128 characters, and the distribution of title lengths is shown in Figure \ref{fig:title-length-distribution}. A word cloud of the most frequent words in titles is shown in Figure~\ref{fig:wordcloud}.


We split the original training data into an 80\% training set (518k images) and a 20\% hold-out validation set (129k images). CLIP embeddings (512-dimensional vectors) are pre-computed for both images and titles to significantly reduce training time.

\subsection{Method}

We employ CLIP \cite{radford_learning_2021}, a vision-language model trained on 400 million image-caption pairs using a contrastive objective. The model consists of an image encoder (\texttt{ViT-B/32}) and a text encoder (Transformer), which project inputs into a shared 512-dimensional embedding space where matching image-text pairs exhibit high cosine similarity. In our framework, we freeze both encoders and use their pre-trained weights without fine-tuning, extracting fixed feature representations for downstream classification.


\subsubsection{Image, text, and location embeddings}  

For each image $x$ with title $t$ and grid centroid coordinates (lat, lon), we extract three types of features:
\begin{align}
z_{\text{img}} &= f_{\text{img}}(x) \in \mathbb{R}^{512} \\
z_{\text{txt}} &= f_{\text{txt}}(t) \in \mathbb{R}^{512} \\
z_{\text{loc}} &= \left[\frac{\text{lat} - 49.9}{61.9 - 49.9}, \frac{\text{lon} - (-8.6)}{2.1 - (-8.6)}\right] \in \mathbb{R}^2
\end{align}


where $f_{\text{img}}$ and $f_{\text{txt}}$ are the frozen CLIP image and text encoders, respectively. Location coordinates are min-max normalised using rough UK geographic boundaries. All features are precomputed and remain fixed during training.

\subsubsection{Multimodal fusion}

We examine how different modalities contribute to classification performance by training models on all possible feature combinations. Table~\ref{tab:modalities} shows the seven variants tested, ranging from single modalities to full multimodal fusion.

\begin{table}[h]
\centering
\begin{tabular}{lcc}
\textbf{Modality combination} & \textbf{Feature vector} & \textbf{Dimension} \\
\hline
Image only & $z_{\text{img}}$ & 512 \\
Title only & $z_{\text{txt}}$ & 512 \\
Location only & $z_{\text{loc}}$ & 2 \\
Image + Title & $[z_{\text{img}}; z_{\text{txt}}]$ & 1024 \\
Image + Location & $[z_{\text{img}}; z_{\text{loc}}]$ & 514 \\
Title + Location & $[z_{\text{txt}}; z_{\text{loc}}]$ & 514 \\
All modalities & $[z_{\text{img}}; z_{\text{txt}}; z_{\text{loc}}]$ & 1026 \\
\bottomrule
\end{tabular}
\caption{Multimodal feature combinations.}
\label{tab:modalities}
\end{table}

\subsubsection{Classification heads}

We experiment with two types of classification heads: a linear layer and a multi-layer perceptron (MLP).

A linear head is applied to the fused embedding \(z \in \mathbb{R}^d\), with weight matrix \(W \in \mathbb{R}^{49 \times d}\) and bias \(b \in \mathbb{R}^{49}\):
\[
  y = Wz + b
\]

The MLP head consists of a hidden layer with ReLU activation and dropout ($p=0.5$), followed by a linear output layer:

\begin{align*}
  y &= W_2\, \mathrm{Dropout}(\mathrm{ReLU}(W_1 z + b_1)) + b_2 \\
  W_1 &\in \mathbb{R}^{256 \times d} \quad
  b_1 \in \mathbb{R}^{256} \quad
  W_2 \in \mathbb{R}^{49 \times 256} \quad
  b_2 \in \mathbb{R}^{49}
\end{align*}

\subsubsection{Training}

We train all models using binary cross-entropy loss with logits (\texttt{BCEWithLogitsLoss}) over a maximum of 200 epochs, with early stopping if validation accuracy does not improve for 10 consecutive epochs. The learning rate follows a cosine annealing schedule with $T_{max}=50$. Training is performed on CPU (MacBook Pro M3) with a batch size of 4096, and most runs complete within 10 minutes depending on input modalities and head architecture.

We run each experiment with and without MixUp (a Beta distribution with \(\alpha = 0.4\)) applied to the fused input vectors to improve generalisation.

At the end of each epoch, we evaluate on the hold-out validation set using subset accuracy, which requires all 49 labels to match exactly. Predictions are computed by applying a sigmoid followed by thresholding at 0.5. We also report macro-averaged $F_1$ score. Model checkpoints are selected based on validation accuracy, which is Kaggle's evaluation metric.

For Kaggle submission, we ensure each image has at least one tag; we do not restrict the maximum number of tags.

\section{Results and Discussion}

Table~\ref{tab:val_results} shows validation and test performance across all input modality combinations, classification head types, and MixUp. Figure~\ref{fig:f1_bar} (Appendix) shows per-class $F_1$ scores on the validation set for our best model, highlighting variation in tag difficulty: \textit{Canals}, \textit{Air transport} or \textit{Railways}, which represent distinct and objective features, achieve high $F_1$ scores; less visually pronounced tags like \textit{Flat landscapes} and \textit{Lowlands} perform poorly. Figure~\ref{fig:misclassified} illustrates the subjectivity of labelling; our model's predicted tags are often as (or even more) appropriate as the original annotations.

\begin{table*}[ht]
  \centering
  \setlength{\tabcolsep}{16pt}
  \begin{tabular}{l|c|cc|cc|cc}
    \toprule
     & & \multicolumn{2}{c|}{\textbf{Val. $F_1$}} 
    & \multicolumn{2}{c|}{\textbf{Val. Acc.}} 
    & \multicolumn{2}{c}{\textbf{Kaggle Acc.}} \\
    
    \textbf{Input Modality} & \textbf{MixUp} & \textbf{Linear} & \textbf{MLP} 
      & \textbf{Linear} & \textbf{MLP} 
      & \textbf{Linear} & \textbf{MLP} \\
    \midrule

    Image & No  & 0.413 & 0.377 & 0.223 & 0.219 & 0.267 & 0.277 \\
    Title & No  & 0.506 & 0.495 & 0.278 & 0.287 & 0.309 & 0.327 \\
    Location & No & 0.000 & 0.000 & 0.000 & 0.000 & -- & -- \\
    Image + Title & No & 0.585 & 0.540 & \textit{0.327} & 0.324 & 0.353 & 0.359 \\
    Image + Location & No & 0.411 & 0.375 & 0.223 & 0.218 & 0.267 & 0.276 \\
    Title + Location & No & 0.504 & 0.494 & 0.276 & 0.287 & 0.307 & 0.326 \\
    Image + Title + Location & No & 0.585 & 0.543 & \textit{0.327} & 0.325 & \textit{0.354} & \textbf{0.360} \\

    \midrule 

    Image & Yes  & 0.432 & 0.376 & 0.227 & 0.217 & 0.264 & 0.273 \\
    Title & Yes  & 0.518 & 0.484 & 0.279 & 0.279 & 0.307 & 0.322 \\
    Location & Yes & 0.000 & 0.000 & 0.000 & 0.000 & -- & -- \\
    Image + Title & Yes & \textit{0.595} & 0.526 & \textit{0.327} & 0.315 & 0.351 & 0.355 \\
    Image + Location & Yes & 0.430 & 0.362 & 0.227 & 0.210 & 0.263 & 0.272 \\
    Title + Location & Yes & 0.518 & 0.491 & 0.279 & 0.284 & 0.307 & 0.325 \\
    Image + Title + Location & Yes & \textbf{0.597} & 0.527 & \textbf{0.328} & 0.314 & 0.350 & 0.354 \\
    
    \bottomrule
  \end{tabular}
  \caption{Performance across input modalities and MixUp settings for linear and MLP heads. Validation accuracy is from the hold-out set; Kaggle accuracy reflects test-set scores from submission. Best values in \textbf{bold}, second-best in \textit{italics}.}
  \label{tab:val_results}
\end{table*}

\paragraph{Adding more modalities generally improves performance}

The best overall results are obtained when combining image, title, and location. While location alone is ineffective (validation scores near zero due to its coarse 1km centroid approximation), it provides small gains when used along with image and text modalities.

\paragraph{Title embeddings are more predictive than image features alone}
This is intuitive; the same contributor typically provides both the title and the tags, making the title a close proxy of the tagging intent. For example, the title-only model achieves a Kaggle accuracy of 0.309 (no MixUp, linear), compared to 0.267 for image-only.

\paragraph{The linear head often matches or outperforms the MLP on validation accuracy} This suggests that the fused CLIP embeddings are already linearly separable enough for this task, and that a more complex classifier does not yield a consistent benefit. However, this trend reverses in Kaggle submissions, where enforcing at least one tag per image appears to favour the MLP.

\paragraph{MixUp provides small gains in validation $F_1$ and accuracy} However, this does not consistently translate to improvements on the Kaggle test set.

\paragraph{Future directions} While we use frozen CLIP embeddings, future work could explore stronger feature extractors such as DINOv2 \cite{oquab_dinov2_2024} for images, SentenceTransformers for text \cite{reimers_sentence-bert_2019, ilyankou_sentence_2024}, and more sophisticated location encoders. Incorporating auxiliary data like land cover, POIs, or satellite imagery along with modality-specific encoders like SatCLIP \cite{klemmer_satclip_2024} or CaLLiPer \cite{wang_multi-modal_2025} may also improve classification. Multimodal LLMs, which already perform well on diverse tasks like road attribute classification \cite{jongwiriyanurak2025vroastvisualroadassessment}, should also be explored.

\paragraph{Dataset value} Beyond the Kaggle challenge, this dataset has potential for applications in tourism, regional image classification, and GeoAI research (particularly in regions lacking street-level imagery). It may support learning regional visual embeddings or classifying landscape types from crowdsourced photographs. Crucially, the task is not to predict objective ground-truth labels, but rather to approximate how a human contributor would describe the image. This introduces subjectivity into the supervision signal, making the dataset well-suited for studying human labelling behaviour and human–AI alignment in multimodal settings.

\section{Conclusion}

We presented a lightweight CLIP-based classifier for multi-label tag prediction of crowdsourced landscape photos from the Geograph dataset. Using frozen CLIP embeddings and a simple trainable head, our model combines image, title, and approximate location inputs. Multimodal fusion improves performance, with text (titles) especially informative, reflecting the contributor-driven nature of the tags. Coarse location data adds some value alongside image and text. Despite the task's subjectivity and strict evaluation, our approach runs efficiently on a laptop and can be generalised to similar multimodal classification tasks. This work supports automated tagging in crowdsourced geographic image archives, enabling better search and discovery across large unlabelled collections.


\begin{acks}
This work was supported by Ordnance Survey \& UKRI Engineering and Physical Sciences Research Council [grant no. EP/Y528651/1].
\end{acks}

\bibliographystyle{ACM-Reference-Format}
\bibliography{articles,manual_refs}


\appendix

\renewcommand{\thefigure}{A.\arabic{figure}}
\setcounter{figure}{0}
\renewcommand{\thetable}{A.\arabic{table}}
\setcounter{table}{0}

\begin{table*}\small
\begin{tabular}{p{2cm} p{4.5cm} p{9.5cm}}
\hline

Group & Tag & Description  \\
\hline

\multirow{6}{*}{Topography}           & Coastal                            & Landforms that occur where the land meets the sea.                                                        \\
                                      & Islands                            & Areas of land completely surrounded by water.                                                             \\
                                      & Flat landscapes                    & Flat landscapes are plains, plateaus, levels, and wide valley floors.                                     \\
                                      & Lowlands                           & Mostly cultivated land at low elevations but not completely flat.                                         \\
                                      & Uplands                            & Mostly uncultivated land at mainly high elevations.                                                       \\
                                      & Geological interest                & Rocks, geology and geophysical processes, above and below ground.                                         \\
                                      \hline

\multirow{11}{*}{\makecell[l]{Natural\\environment}} & Air, Sky, Weather                  & Air as habitat; weather as an agent of geophysical processes or typical   of a place.                     \\
                                      & Estuary, Marine                    & Tidal, salt-water habitats and environs.                                                                  \\
                                      & Lakes, Wetland, Bog                & Natural and man-made waterbodies and their margins.                                                       \\
                                      & Rivers, Streams, Drainage          & Natural and man-made watercourses.                                                                        \\
                                      & Grassland                          & Landscapes and habitats which are predominantly and permanently grass.                                    \\
                                      & Rocks, Scree, Cliffs               & Exposed rock of any size.                                                                                 \\
                                      & Barren Plateaux                    & Landscapes supporting little vegetation.                                                                  \\
                                      & Moorland                           & Landscapes on generally wet ground with heather or long, rough grass   species.                           \\
                                      & Heath, Scrub                       & Landscapes dominated by heather and ling on mostly dry ground.                                            \\
                                      & Woodland, Forest                   & Natural and planted woods, managed or not.                                                                \\
                                      & Wild Animals, Plants and Mushrooms & Choose the category of Natural environment in which the wildlife was   photographed.                      \\
                                      \hline

\multirow{12}{*}{Human use}           & Farm, Fishery, Market Gardening    & Growing crops outdoors or under cover. Rearing animals and fish.                                          \\
                                      & Quarrying, Mining                  & Extraction of stone, clay, coal and other fuels, minerals, ores, and   aggregates.                        \\
                                      & Water resources                    & The water cycle from river or ground to tap.                                                              \\
                                      & Energy infrastructure              & Generation and distribution of energy.                                                                    \\
                                      & Country estates                    & Big houses, extensive grounds, policies and parkland. Usually private.                                    \\
                                      & Industry                           & Any manufacturing, processing or maintenance activity not covered by the   other `human use' categories.  \\
                                      & Defence, Military                  & Any current military and some historic sites.                                                             \\
                                      & Construction, Development          & Constructing or converting, maintaining or demolishing structures of any   kind.                          \\
                                      & Business, Retail, Services         & Shops, superstores, offices, depots and other business premises.                                          \\
                                      & Sport, Leisure                     & Any sport or leisure activity and facilities for it.                                                      \\
                                      & Waste, Waste management            & Facilities for dealing with domestic refuse, recycling, industrial waste,   sewage.                       \\
                                      & Derelict, Disused                  & Abandoned, disused, or ruined places.                                                                     \\
                                      \hline

\multirow{13}{*}{Human habitat}       & City, Town centre                  & Town and city streets, squares, townscapes, scenes.                                                       \\
                                      & Suburb, Urban fringe               & Outer areas of towns and cities and the edge of adjoining countryside.                                    \\
                                      & Village, Rural settlement          & Villages, hamlets, farmhouses, barn conversions, isolated dwellings.                                      \\
                                      & Park and Public Gardens            & Including playgrounds, village greens.                                                                    \\
                                      & Public buildings and spaces        & Local and central government offices; pedestrian areas; public and   community buildings.                 \\
                                      & Housing, Dwellings                 & Houses, cottages, apartments and flats.                                                                   \\
                                      & Educational sites                  & Public and private places for learning and training of all kinds for all   ages.                          \\
                                      & Health and social services         & Places providing treatment, care or promoting health and wellbeing.                                       \\
                                      & Historic sites and artefacts       & Any place, object or activity designated or claiming to be historic.                                      \\
                                      & Religious sites                    & Buildings and land, old and new, associated with any faith.                                               \\
                                      & Boundary, Barrier                  & Physical boundaries and barriers. Legal and administrative boundaries,   marked or not.                   \\
                                      & People, Events                     & People and activities rather than the place they are in.                                                  \\
                                      & Burial ground, Crematorium         & Land and buildings devoted to the disposal of the dead.                                                   \\
                                      \hline

\multirow{7}{*}{Communications}       & Canals                             & Navigable canals and rivers. Not land drainage.                                                           \\
                                      & Docks, Harbours                    & On the coast and in estuaries. Ferries. Fish. Freight. Facilities.                                        \\
                                      & Railways                           & Includes off-street urban light rail and older industrial tramroads. Any   locomotives and rolling stock. \\
                                      & Paths                              & Any way that you can walk or ride along that isn't a driveable road or   track.                           \\
                                      & Roads, Road transport              & Any way that can be driven or ridden on. Ancillary places and equipment.   Vehicles.                      \\
                                      & Air transport                      & Airports, airfields, helipads and associated buildings and equipment.                                     \\
                                      & Communications                     & Any means of sending signals and messages over long distances.   \\     
\hline

\end{tabular}

\caption{A list of 49 tags and descriptions used to label images in the Kaggle Geograph challenge.}

\label{tab:tags}

\end{table*}

\begin{figure*}[ht]
    \centering
    \includegraphics[width=\linewidth]{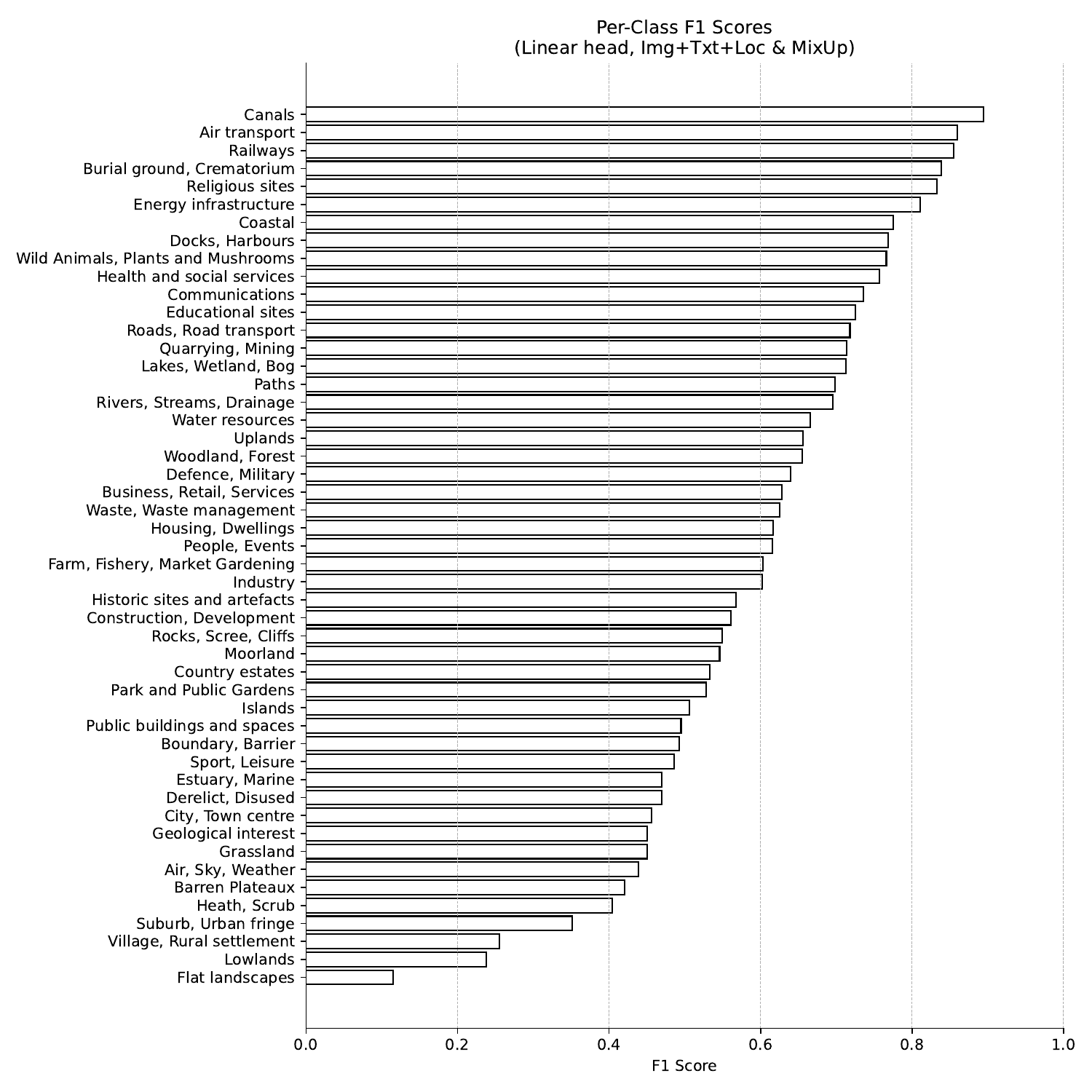}
    \caption{Per-class $F_1$ scores for the best performing model on the validation set.} 
    \label{fig:f1_bar}
\end{figure*}

\begin{figure*}[ht]
    \centering
    \includegraphics[width=\linewidth]{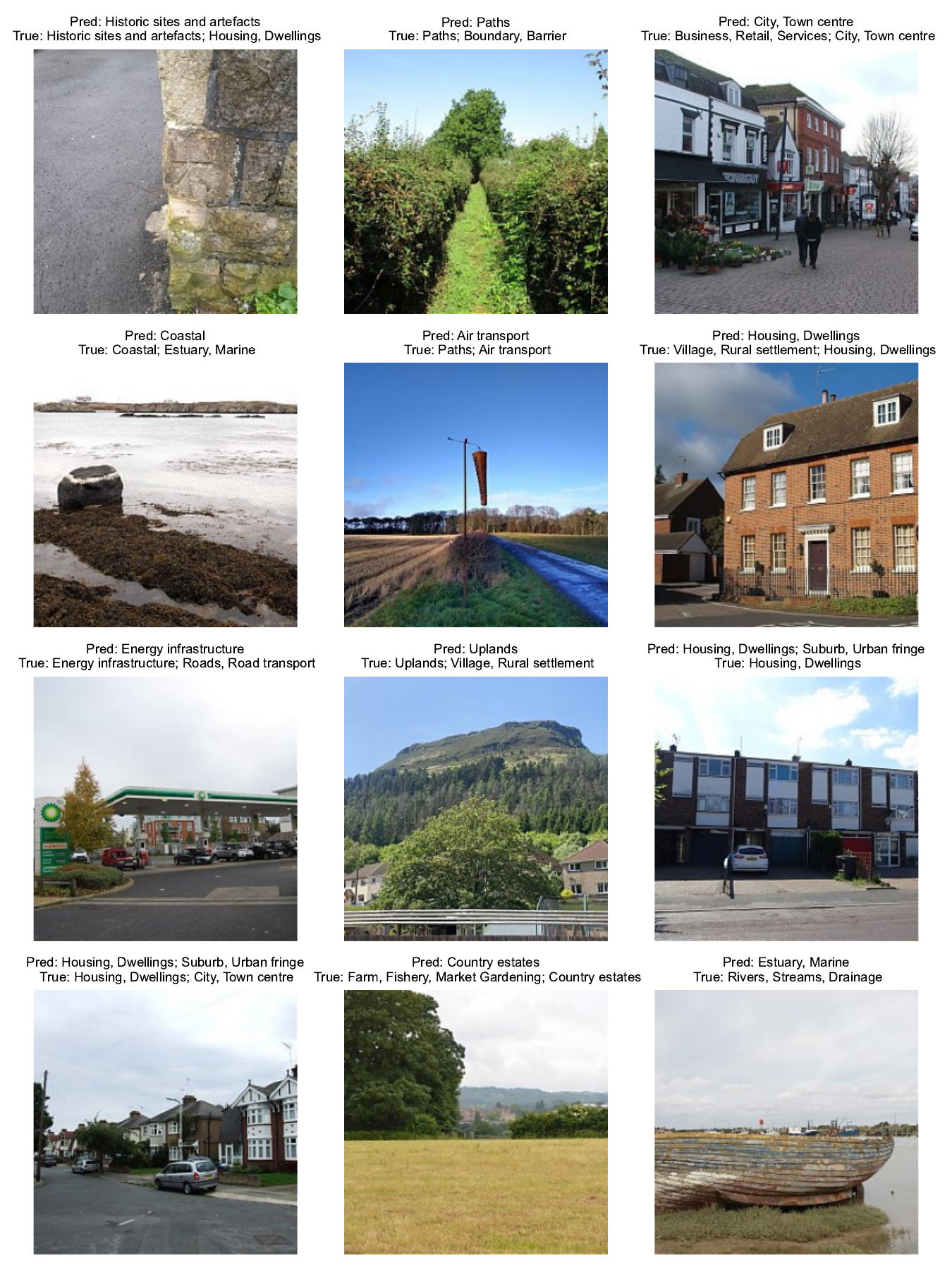}
    \caption{Validation examples where predicted tags differ from original labels, often due to incomplete annotations.} 
    \label{fig:misclassified}
\end{figure*}

\end{document}